\documentclass[conference]{IEEEtran}
\IEEEoverridecommandlockouts
\usepackage{cite}
\usepackage{amsmath,amssymb,amsfonts}
\usepackage{algorithmic}
\usepackage{graphicx}
\usepackage{textcomp}
\usepackage{bm}
\usepackage{multirow}
\usepackage{booktabs}
\usepackage{color,xcolor}

\def\BibTeX{{\rm B\kern-.05em{\sc i\kern-.025em b}\kern-.08em
    T\kern-.1667em\lower.7ex\hbox{E}\kern-.125emX}}
\begin{document}

\title{Adaptive Temporal Motion Guided Graph Convolution Network for Micro-expression Recognition
}

\author{\IEEEauthorblockN{Fengyuan Zhang}
\IEEEauthorblockA{
\textit{Renmin University of China}\\
fy.zhang@ruc.edu.cn}
\\
\IEEEauthorblockN{Xinjie Zhang}
\IEEEauthorblockA{
\textit{Renmin University of China}\\
zhangxinjie827@ruc.edu.cn}
\and
\IEEEauthorblockN{Zhaopei Huang}
\IEEEauthorblockA{
\textit{Renmin University of China}\\
huangzhaopei@ruc.edu.cn}
\\
\IEEEauthorblockN{Qin Jin*
\thanks{*Corresponding author. }}
\IEEEauthorblockA{
\textit{Renmin University of China}\\
qjin@ruc.edu.cn}
}

\maketitle

\begin{abstract}
Micro-expressions serve as essential cues for understanding individuals' genuine emotional states. Recognizing micro-expressions attracts increasing research attention due to its various applications in fields such as business negotiation and psychotherapy. However, the intricate and transient nature of micro-expressions poses a significant challenge to their accurate recognition. Most existing works either neglect temporal dependencies or suffer from redundancy issues in clip-level recognition. In this work, we propose a novel framework for micro-expression recognition, named the Adaptive Temporal Motion Guided Graph Convolution Network (\textbf{ATM-GCN}). Our framework excels at capturing temporal dependencies between frames across the entire clip, thereby enhancing micro-expression recognition at the clip level. Specifically, the integration of Adaptive Temporal Motion layers empowers our method to aggregate global and local motion features inherent in micro-expressions. Experimental results demonstrate that ATM-GCN not only surpasses existing state-of-the-art methods, particularly on the Composite dataset, but also achieves superior performance on the latest micro-expression dataset CAS(ME)$^3$.
\end{abstract}


\section{Introduction}
Facial expressions are conscious human reactions to certain stimuli and play critical roles in our daily communications. 
Many studies~\cite{Cai2018, Ding2017} have explored the complex mechanisms of facial expressions and further attempted to understand the underlying emotional cues behind them.
These efforts are primarily made in macro-expressions that are often easily noticeable by individuals. Micro-expressions are another form of facial expression that appear for a fleeting duration of less than 0.5 seconds. Although micro-expressions are very brief, they serve as essential cues for understanding individuals’ genuine emotional states and often appear when people are trying to hide their feelings. Automatic micro-expression recognition (MER) has attracted increasing research attention due to its diverse applications in fields such as business negotiation and psychotherapy etc.

\begin{figure}[htb]
\centering
\includegraphics[width=0.8\columnwidth]{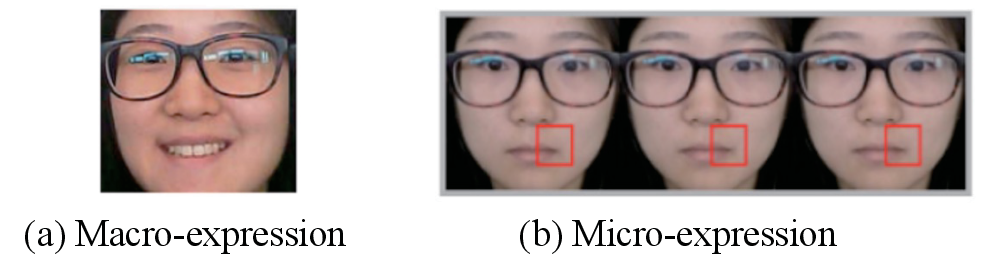}
\vspace{-8pt}
\caption{Macro- vs Micro-expression of Happiness}
\label{fig:expression}
\end{figure}

However, the accurate recognition of micro-expressions still remains a formidable challenge. 
First, unlike macro-expressions, micro-expression frames actually share stronger temporal dependencies, so it is difficult to accurately recognize micro-expressions through one single frame, as shown in Figure \ref{fig:expression}. 
Second, micro-expressions are complex combinations of several subtle facial movements. Therefore, effectively modeling the relationship between these facial movements that make up a certain expression is very important. 

\begin{figure*}[ht]
\centering
\includegraphics[width=2\columnwidth]{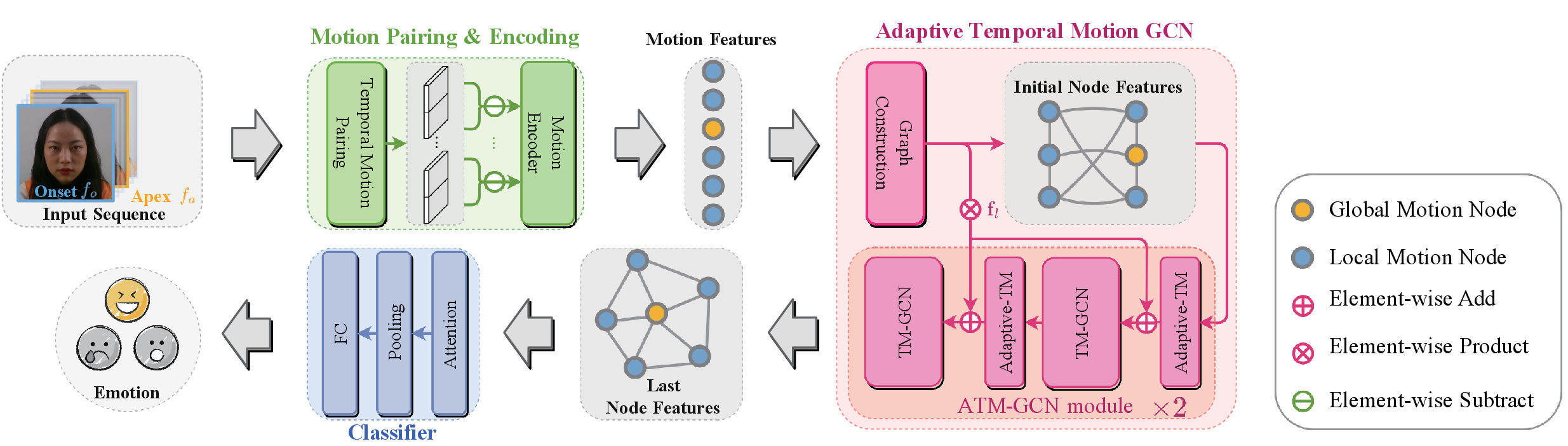} 
\caption{An overview of the proposed ATM-GCN approach for micro-expression recognition. $f_o$ and $f_a$ represent the Onset and Apex frame respectively for simplification. The input sequence is first input into the Motion Pairing \& Encoding module for extracting motion features between frame pairs, which are then aggregated through the Adaptive Temporal Motion GCN (ATM-GCN) module. Finally, a Classifier module is utilized to get the predicted micro-expression for the input sequence.}
\label{fig:atm-gcn}
\end{figure*}

Over the years, different attempts have been made to address these challenges. 
Some methods such as \cite{Li2022f} only use the Onset frame and the Apex frame in a clip and extract the micro-expression features inherent therein, which helps capture the most important motion in a clip, but it discards critical temporal information. 
On the other hand, methods like SLSTT \cite{Zhang2022b} end up encountering potential information redundancy when utilizing full frame sequences. 
Inspired by GCN, some works \cite{Xie2020} construct graphs by learning the relation between different facial movements. Although more theoretically explainable, these works require extra labels for training. Moreover, they neglect the temporal information, resulting in insufficient utilization of the entire clip. 

In this work we propose a new GCN-based MER method, named Adaptive Temporal Motion guided Graph Convolution Network (\textbf{ATM-GCN}).
With the full frame sequence as input, our method manages to pay more attention to the most important motion information from the whole sequence. 
Moreover, our method mitigates redundancy by modeling the relation of motion information in different temporal locations and further modifies the relation adaptively during training. The main contributions of this work include: 1) We propose the Adaptive Temporal Motion guided Graph Convolutional Network (\textbf{ATM-GCN}) that incorporates temporal dependencies of frame pairs to fully exploit motion information for MER; 2) To dynamically adjust the transmission of motion information between nodes, a new Adaptive Temporal Motion (\textbf{Adaptive-TM}) layer is specifically designed to modify the importance of different nodes during training; 3) Experimental results demonstrate that our proposed ATM-GCN outperforms existing state-of-the-art methods on both Composite and CAS(ME)$^3$ datasets.

\section{Related Works}

\subsection{Micro-Expression Recognition}
Earlier works for MER largely rely on traditional approaches that extract handcrafted features. Despite their contributions, these methods primarily process signals at a low level, often neglecting crucial semantic information. In recent years, deep learning methods have demonstrated superior performance, with an enhanced ability to learn embedded semantics within input signals. 
The existing approaches for MER can be roughly grouped into three categories: frame-based, sequence-based, and multi-modal approaches.

Some frame-based methods access only the most significant frames within a clip: the Onset frame (the start of a micro-expression where the face transitions from a neutral expression) and the Apex frame (the peak of a micro-expression where the expression is most intense). Although such approaches minimize the complexity of the input data and capture the most crucial motion information. 
they discard most of the temporal dependencies within a sequence. 
Different from frame-based methods, sequence-based methods use full sequences as input, aiming to preserve comprehensive temporal information. 
Although sequence-based approaches appear theoretically more effective, their state-of-the-art cannot be compared with that of frame-based approaches because sequential data may suffer from temporal redundancy due to the possible similarity between neighboring frames. A model might learn some irrelevant information like gender or color space distribution of a given clip. 


\subsection{GCNs for MER}

Graph Convolutional Networks (GCN), a novel paradigm introduced by \cite{Kipf2017a} for processing graph-structured data, have gained prominence in recent years. GCN operates in the spectral domain to extract high-level global patterns from graphs.
Recently, researchers have attempted to leverage GCN for MER. It is clear that facial micro-expressions naturally form topological graphs by interconnecting important facial regions (e.g., eye, nose, mouth).
However, current GCNs for MER lack robust modeling of temporal dependencies, which is crucial to MER. And most GCN-based MER methods concentrate more on the relationships between different parts within a static frame, but less on the dynamic motion feature between frames. Therefore, the temporal transitions between frames should be explored in GCNs to ensure a comprehensive understanding of micro-expressions.

\section{Method}


Figure \ref{fig:atm-gcn} illustrates our proposed ATM-GCN framework which contains three key modules: Temporal Motion Pairing \& Encoding, Adaptive Temporal Motion guided GCN, and Classifier. 

\subsection{Temporal Motion Pairing \& Encoding}

We take the whole micro-expression frame sequence $F = \{f_i\}_{i=1}^{L}\in \mathbb{R}^{L \times C \times H \times W}$ as input, which contains $L$ consecutive RGB frames $f_i$. 
Specifically, there are two special frames in a micro-expression sequence: 1) the Onset frame, referring to the beginning frame of a micro-expression, and 2) the Apex frame, referring to the peak of a micro-expression. We denote the two special frame as $f_{onset}$ and $f_{apex}$. Normally the index of the Onset frame is set to $onset=1$ as it is the first frame of a micro-expression clip.

\begin{figure}[t]
\centering
\includegraphics[width=1\columnwidth]{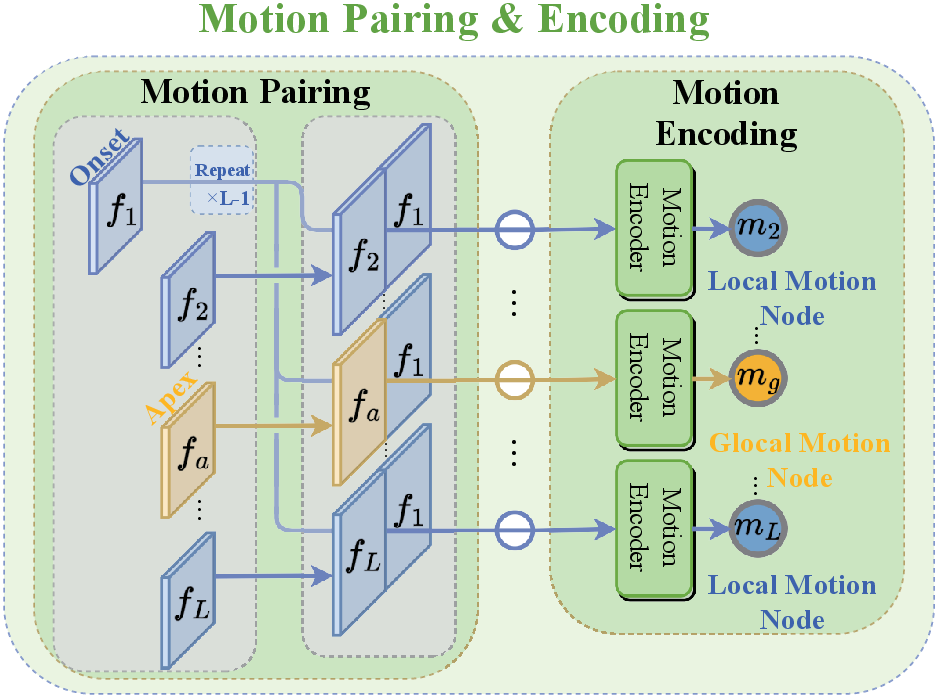} 
\caption{Detailed illustration of our Motion Pairing \& Encoding module. Frame $f_a$ denotes the Apex frame $f_{apex}$  and $m_a$ denotes the corresponding $m_{apex}$ for simplicity. The Onset frame is paired with each of other $L$-1 frames. We then extract motion features from the pairs for nodes initialization.}
\label{fig:pairing}
\end{figure}

We construct multiple frame pairs along the temporal dimension to adaptively capture facial movements through the whole sequence. 
As shown in Figure \ref{fig:pairing}, given a micro-expression sequence $F$, we design the \textbf{Temporal Motion Pairing} module to produce frame pairs $P$ as follows:

\begin{equation}
P = \{p_i\}_{i=2}^{L}=\{(f_{i}, f_{onset})\}_{i=2}^{L}
\label{eq:pairing}
\end{equation}

Based on the frame pairs, we further compute the frame differences, which are fed into the Motion Encoder. Specifically, for frame pairs $(f_{i}, f_{onset})$, we produce the frame difference as ${\Delta f}_{i}= f_{i} - f_{onset}$.
We thus form the inputs $\{{\Delta f}_{i}\}_{i=2}^{L}$ for the Motion Encoder, which subsequently outputs the motion feature $m_i \in \mathbb{R}^{N_m}$, where $N_m$ denotes the size of motion features. Extracting robust motion features is crucial for our ATM-GCN. Inspired by \cite{Li2022f}, we adopt the \textbf{Continuous Attention Block} (CA block) in the Motion Encoder:
\begin{equation}
m_i = \mathcal{M}({\Delta f}_{i},\mathcal{T}(i))
\label{eq:motion_encoder}
\end{equation}
where $\mathcal{M}$ is our Motion Encoder, $i$ is the frame index of $f_i$, and $\mathcal{T}$ refers to temporal positional encoding. The $i$ can also reflect the temporal distance between the current frame $f_i$ and the Onset frame $f_{onset}$. So we can explicitly inject the temporal distance information of the current pair by adding a $\mathcal{T}(i)$ into the input.

In summary, through the Temporal Motion Pairing \& Encoding module, we produce the motion features $\{m_i\}_{i=2}^{L}$, adaptively focusing on facial movements throughout the whole sequence.

\subsection{Adaptive Temporal Motion guided GCN}

\begin{figure}[htb]
\centering
\includegraphics[width=1\columnwidth]{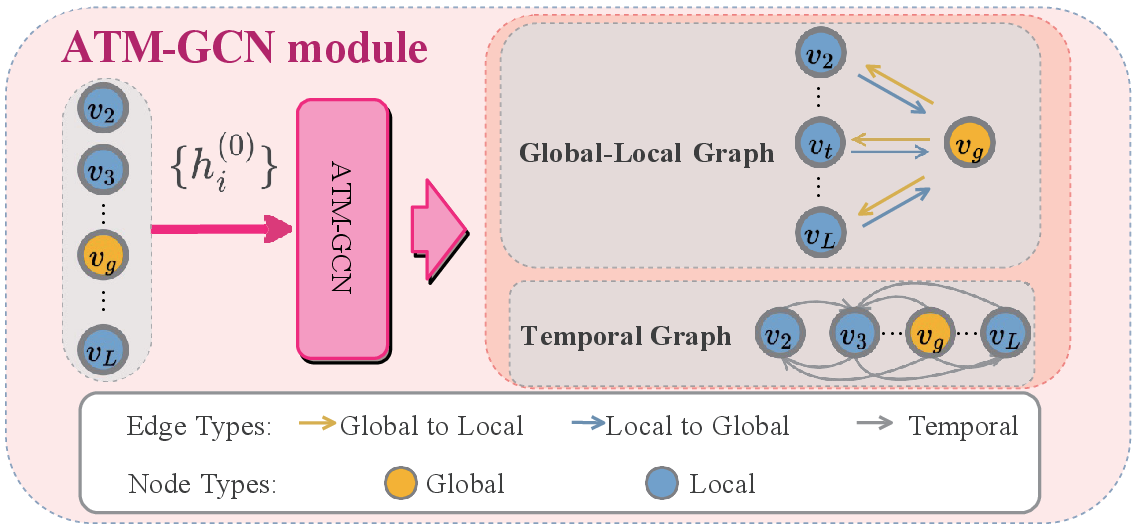} 
\caption{The detailed graph construction process of our ATM-GCN. 
$v_g$ represents the Global Motion Node $v_{global}$ for simplification. $v_1$ is trivial and removed in the graph construction process. The initial node features $\{h_i^{(0)}\}$ are input into the ATM-GCN module for graph construction and processing.}
\label{fig:graph}
\end{figure}

We design a GCN based architecture which specifically emphasizes the temporal dimension of micro-expressions, named Adaptive Temporal Motion guided GCN \textbf{(ATM-GCN)}.
Figure \ref{fig:graph} illustrates the detailed graph construction process of our ATM-GCN module.

\subsubsection{Nodes}

We denote each node in our graph as $\{v_i\}_{i=2}^{L}$ and initialize it by the corresponding output from the Motion Encoder, formulated as $h_i^{(0)}=m_i$,
where $h_i^{(0)}$ with a superscript $(0)$ denotes the initial node features.

Consequently, there are two types of nodes in our graph: the Global Motion Node and Local Motion Nodes. We set one Global Motion Node $v_{global}$ for a micro-expression sequence, which is initialized by $m_{apex}$, the motion feature between the Onset and Apex frames. We consider ${v}_{global}$ captures relatively global motion information of the whole sequence. We set all other nodes as Local Motion Nodes $v_i \in \mathcal{V}_{local}$, which capture relatively local motion information which may also be important for micro-expression recognition.

\subsubsection{Edges}

We first build edges to capture temporal dependencies among all nodes in our graph. 
Theoretically, each node should interact with all the other nodes in the sequence. But that would result in very dense graph, and the model would encounter overfitting issues due to the small sample-size. Therefore, 
we fix the window size to $2w$, so that each node $v_i$ only interacts with nodes within the window:
\begin{equation}
\begin{aligned}
    \mathcal{E}_i = \{(v_i,v_j)\}_{j=\max(2,i-w)}^{\min(L,i+w)}
\end{aligned}
\end{equation}
where $\mathcal{E}_i$ is the edge set for node $v_i$ and $i\in[2,L],i \neq global$. All the edges in $\mathcal{E}_i$ are directed edges, and $(v_i,v_j)$ denotes the edge from node $v_i$ to node $v_j$.

Furthermore, we also build edges between two types of nodes to enhance global and local feature aggregation. Specifically, for $v_i \in \mathcal{V}_{local}$, we construct extra edges as shown in Equation (\ref{eq:local_nodes}):
\begin{equation}
\label{eq:local_nodes}
\widetilde{\mathcal{E}}_i = \mathcal{E}_i \cup \{(v_{i},v_{global})\}
\end{equation}
where $\widetilde{\mathcal{E}}_{i}$ is the edge set for $v_i \in \mathcal{V}_{local}$. And for $v_{global}$, it is connected to all the other nodes in the sequence:
\begin{equation}
\label{eq:global_nodes}
\widetilde{\mathcal{E}}_{global} = \mathcal{E}_{global} \cup \{(v_{global},v_{i})\}
\end{equation}
where $\widetilde{\mathcal{E}}_{global}$ is the edge set for $v_{global}$. 
By now we have built two types of edges for our graph that can capture both temporal dependencies and important local features for micro-expression recognition.

\subsubsection{Edge Weighting Strategies}


We first set the initial weight for $e_{ij}$ based on similarity between $v_i$ and $v_j$, where $e_{ij}$ is the directed edge $(v_i,v_j) \in \widetilde{\mathcal{E}}_{i}$ for simplification. We assume that if two nodes are similar, they both contain similar micro-expression pattern. Following \cite{Skianis2018}, we use the angular similarity to capture the similarities between nodes:
\begin{equation}
\mathcal{A}_{ij}^{(l)}=1-\frac{\arccos(sim(h_i^{(l)},h_j^{(l)}))}{\pi}
\end{equation}
where $\mathcal{A}^{(l)}$ is the adjacent matrix of $l$-th GCN layer and $\mathcal{A}_{ij}^{(l)}$ denotes the corresponding edge weight for $e_{ij}$. $h_i^{(l)}$ and $h_j^{(l)}$ denote the node features of $l$-th GCN layer.

Besides, we adjust the edge weights based on temporal distance between nodes. We think if two nodes are temporally more distant from each other, their interactions should also be weaker. Therefore, we add extra decaying factor for each edge weight as shown below:
\begin{equation}
\mathcal{A}_{ij}^{(l)}=(1-\frac{\arccos(sim(h_i^{(l)},h_j^{(l)}))}\pi) \times \exp{\frac{-|i-j|}{\tau}}
\end{equation}
where $\tau$ is used to control the decaying rate, and set to $10$ in implementation. By applying the decaying factor in the edge weights, we can further enhance temporal dependencies in our graph.

Moreover, we further add reweighting factors based on node types. We believe that the node feature of $v_{global}$ is more important and should have stronger interactions with other nodes. Therefore, to enhance the global interactions, we add a reweighting factor for each $\mathcal{A}_{ij}^{(l)}$ as follows:
\begin{equation}
\label{eq:reweighting}
\widetilde{\mathcal{A}}_{ij}^{(l)}
=\left\{
\begin{aligned}
\lambda_{local} \mathcal{A}_{ij}^{(l)},&e_{ij} \in \widetilde{\mathcal{E}}_i\\
\lambda_{global} \mathcal{A}_{ij}^{(l)},&e_{ij} \in \widetilde{\mathcal{E}}_{global}
\end{aligned}
\right.
\end{equation}
where $\lambda_{local}$ and $\lambda_{global}$ are reweighting factors. We make sure $\lambda_{local} < \lambda_{global}$ to enhance global interactions.

\subsubsection{Adaptive Temporal Motion Layer}
We further design and integrate Adaptive Temporal Motion (\textbf{Adaptive-TM}) layers in our ATM-GCN module to adaptively aggregate global and local motion features inherent in micro-expressions. We denote $G^{(l)}$ as our  $l$-th Temporal Motion GCN (\textbf{TM-GCN}) layer which can be formulated as:
\begin{equation}
    h_i^{(l)} = G^{(l)}(h_i^{(l-1)},\widetilde{\mathcal{A}}^{(l-1)})
\end{equation}
where $h_i^{(l)}$ and $h_i^{(l-1)}$ denotes the node features of $l$-th and ($l$-1)-th TM-GCN layer, and $\widetilde{\mathcal{A}}^{(l-1)}$ denotes the adjacent matrix of ($l$-1)-th layer. For $l$=0, $h_i^{(0)}$ is the initial node features.

In most GCN-based methods, the adjacent matrix is fixed in all GCN layers. Here we integrate Adaptive-TM layer before TM-GCN layers. Specifically, for $l$-th TM-GCN layer $G^{(l)}$, we adjust the adjacent matrix $\widetilde{\mathcal{A}}^{(l)}$ formulated as:
\begin{equation}
\label{eq:amd}
    \widetilde{\mathcal{A}}^{(l)} =  FC(\widetilde{\mathcal{A}}^{(l-1)}) \times (1-\mathrm{f}_{l}) + \widetilde{\mathcal{A}}^{(0)} \times \mathrm{f}_{l}
\end{equation}
where $\mathrm{f}_{l}$ is the forgetting rate. We think the original adjacent matrix preserves important initial node features, therefore we adjust the adjacent matrix by $\mathrm{f}_l$ in a weighted moving average manner during inference.

\subsection{Classifier}

The output of the last TM-GCN layer $\{h_i^{(N_{\mathcal{G}})}\}_{i=2}^{L}$ is utilized for classification, where $N_{\mathcal{G}}$ denotes the last TM-GCN layer. We feed the last node features into Self-Attention layer for feature aggregation, followed by a pooling layer and a fully-connected layer to estimate classification probabilities:
\begin{equation}
\bm{\hat{y}} = FC(Pooling(Attention(\{h_i^{(N_{\mathcal{G}})}\}_{i=2}^{L})))
\label{eq:classifier}
\end{equation}
where $\bm{\hat{y}}\in \mathbb{R}^{c}$ is the estimated probabilities and $c$ is the number of micro-expressions in the specific task.

\section{Experiment}

\subsection{Experiment Settings}

We conduct experiment on the following representative micro-expression datasets:

\noindent \textbf{SAMM} \cite{Davison2018b} contains 159 micro-expression clips from 32 participants, with a resolution of $2040 \times 1088$ and the facial area approximately $400 \times 400$ at 200 frames per second (FPS). SAMM originally includes 8 emotion classes, and normally only 5 classes are utilized for MER: Anger, Happiness, Surprise, Sadness and Others.

\noindent \textbf{CASMEII} \cite{Yan2014} contains 256 micro-expression videos from 26 subjects with a cropped size of $280 \times 340$ at 200 FPS. The original CASMEII includes 7 emotion classes, and only 5 classes are used: Happiness, Disgust, Repression, Surprise, and Others.

\noindent \textbf{Composite} \cite{See2019} is proposed by MEGC2019, in which all ME samples from CASME II, SAMM and SMIC \cite{Li2013} were combined into a single composite dataset. This can provide a more realistic scenario where ME samples are captured from different environments with more subjects.

\noindent \textbf{CAS(ME)$^3$} \cite{Li2022e} is the latest dataset for micro-expression analysis. It provides 1,109 labeled micro-expressions and 3,490 labeled macro-expressions, which is larger than all of the previous micro-expression datasets and can provide relatively abundant samples for training deep methods. The annotations of each expression include 7 emotion classes and corresponding Facial Action Units (FAU).

Following previous works, we adopt Accuracy (ACC), unweighted F1 score (UF1) and unweighted Average Recall (UAR) as evaluation metrics:
\begin{equation}
 \mathrm{UF1}=\frac{1}{C} \sum_{i=1}^{C} \frac{2 \times \mathrm{TP}_{i}}{\mathrm{TP}_{i}+\mathrm{FP}_{i}+\mathrm{FN}_{i}} \quad \mathrm{UAR}=\frac{1}{C} \sum_{i=1}^{C} \frac{\mathrm{TP}_{i}}{N_{i}}
\label{eq:uf1-uar}
\end{equation}

Leave One Subject Out (LOSO) is used as the validation protocol for MER, which leaves out all samples of one subject for model performance evaluation. The overall performance is evaluated by averaging the performances of all subjects.

\subsection{Implementation Details}


\subsubsection{Model Configuration.}
We use 4 CA blocks and 4 ATM-GCN layers as experiment settings for bigger dataset CAS(ME)$^3$. The window size is set as $w=2$ by default, forcing ATM-GCN to learn dense temporal dependencies between motions. For datasests like SAMM and CASMEII, we reconfigure our model in a way more suitable to small datasets. We use 2 CA blocks and 2 ATM-GCN layers and downsample the input clip to 30Hz to reduce complexity. Moreover, we set the window size as $w=1$ to learn coarser temporal dependencies for small datasets.


\subsubsection{Training Details.}

For all datasets, we resize the input frames to $224\times224$. For small datasets like SAMM and CASMEII, we adopt random cropping and color jittering to augment the training data. At the training stage, we adopt AdamW to optimize our model, with learning rate initialized to $0.0001$, decreasing at an exponential rate in 50 epochs. The batch size is set to 16 per GPU, and we use  $4\times40$GB A100 to train our model for all settings. To handle unbalanced conditional distribution problem in micro-expression datasets, we adopt Focal loss for optimization.

\begin{table}[t]
\centering 
  \caption{Performance comparison on CAS(ME)$^3$.}
  \label{tab:casme3_result}
\begin{tabular}{c|cc|cc}
\toprule
\multirow{2}{*}{Model}           & \multicolumn{2}{|c|}{7-class} & \multicolumn{2}{|c}{4-class} \\ & UF1 & UAR  & UF1 & UAR
\\ \midrule
Baseline(RGB) \cite{Li2022e}     & 17.59  & 18.01  & 29.15 & 29.10
\\
Baseline(RGB-D) \cite{Li2022e}     & 17.73  & 18.29  & 30.01 & 29.82
\\
$\mu$-BERT \cite{Nguyen2023}     & 32.64  & 32.54  & 47.18 & 49.13
\\
\midrule
\textbf{ATM-GCN}     & \textbf{43.08}  & \textbf{42.83}  & \textbf{54.23} & \textbf{53.49}
\\ \bottomrule
\end{tabular}
\end{table}

\begin{table}[t]
\centering
  \caption{Performance comparison on Composite dataset}
  \label{tab:composite_class3_result}
\begin{tabular}{c|ccc}
\toprule
\multirow{1}{*}{Model}  
& UF1 & UAR\\
\midrule
MTMNet  & 86.40 & 85.70\\
NMER  & 78.85 & 78.24 \\
GRAPH-AU  & 79.14 & 79.33 \\
ICE-GAN  & 84.50 & 84.10 \\
BDCNN  & 85.09 & 85.00 \\
MiMaNet  & 88.30 & 87.60 \\
$\mu$-BERT  & 89.03 & 88.42 \\
\midrule
\textbf{ATM-GCN} & \textbf{91.20} & \textbf{90.22}
\\ \bottomrule
\end{tabular}
\end{table}

\subsection{Experiment Results}

Table \ref{tab:casme3_result} presents the performance comparison between our ATM-GCN and the baseline method \cite{Li2022e} and the latest method $\mu$-BERT \cite{Nguyen2023} on the CAS(ME)$^3$ dataset. Here baseline(RGB) uses RGB frames as input, and baseline(RGB-D) adds additional depth information as extra channel to perform multi-modal MER. 
Our ATM-GCN significantly outperforms both baseline(RGB) and baseline(RGB-D) for both 7-class and 4-class recognition. Compared to $\mu$-BERT, the state-of-the-art method on CAS(ME)$^3$, we achieve better results in both UF1 and UAR for all the settings. Moreover, as shown in Table \ref{tab:composite_class3_result}, our ATM-GCN outperforms other methods by at least +2\% on the Composite dataset as well. 

\begin{table}[t]
\centering
  \caption{Performance comparison on CASMEII and SAMM}
  \label{tab:casme2_samm_class3_result}
\begin{tabular}{c|cc|cc}
\toprule
\multirow{2}{*}{Model}           & \multicolumn{2}{|c|}{CASMEII} & \multicolumn{2}{|c}{SAMM}\\ & UF1 & ACC & UF1 & ACC
\\ \midrule
OFF-ApexNet \cite{Gan2019a}    & 86.97  &  88.28 & 54.23 &  68.18
\\
STSTNet \cite{Liong2019a}    & 83.82  &  86.86 & 65.18 &  68.10
\\
AU-GACN \cite{Xie2020}    & 35.50  &  71.20 & 43.30 &  70.20 
\\
MTMNet \cite{Xia2020b}     & 70.10  &  75.60 & 73.60 &  74.10 
\\
MiMaNet \cite{Xia2021a}  & 75.90  &  79.90 & 76.40 &  76.70 
\\
$\mu$-BERT \cite{Nguyen2023} & 90.34 & 89.14 & \textbf{83.86} & \textbf{84.75}
\\
\midrule
\textbf{ATM-GCN}     & \textbf{90.48}  & \textbf{90.42} & 79.20 & 80.49
\\ \bottomrule
\end{tabular}
\end{table}

We also validate ATM-GCN on smaller datasets such as SAMM and CASMEII. We compare our methods with some latest frame-based and sequence-based MER methods. Results of 3-class recognition are shown in Table \ref{tab:casme2_samm_class3_result}.
Our ATM-GCN is comparable to previous MER methods. Compared with previous works MTMNet~\cite{Xia2020b} and MiMaNet~\cite{Xia2021a}, our method achieves more than +10\% improvement on both UF1 and ACC, and outperforms GCN-based methods such as AU-GACN~\cite{Xie2020}.
Although our ATM-GCN under-performs $\mu$-BERT on SAMM, it is worth noting that $\mu$-BERT requires extra self-supervised training stage to learn from large amounts of unlabeled data.

\begin{table}[t]
\begin{center}
  \caption{Ablation study on CAS(ME)$^3$.}
  \label{tab:ablation}
\begin{tabular}{c|cc|cc}
\toprule
\multirow{2}{*}{Model}           & \multicolumn{2}{|c|}{7-class} & \multicolumn{2}{|c}{4-class} \\ & UF1 & UAR & UF1 & UAR
\\ \midrule
ATM-GCN(\textit{w/o} GCN)     & 28.88  & 28.42 & 41.10 & 42.15
\\
ATM-GCN(\textit{w/o} motion)     & 37.37  & 37.19 & 49.01 & 50.52
\\
ATM-GCN(\textit{w/o} ATM)     & 40.39  & 40.12 & 52.20 & 52.79
\\
\midrule
\textbf{ATM-GCN(full)}     & \textbf{43.08}  & \textbf{42.83} & \textbf{54.23} & \textbf{53.49}
\\ \bottomrule
\end{tabular}
\end{center}
\end{table}

\subsection{Ablation Study}

We carry out experiments to ablate different components of our model. We design 4 variants of ATM-GCN: ATM-GCN(\textit{w/o} GCN), ATM-GCN(\textit{w/o} motion), ATM-GCN(\textit{w/o} ATM) and ATM-GCN(full). In ATM-GCN(\textit{w/o} GCN), we remove GCN module and directly use motion features $\{m_i\}$ for classification. In ATM-GCN(\textit{w/o} motion), we ignore the Temporal Motion Pairing stage and directly extract features from sequence frames. In ATM-GCN(\textit{w/o} ATM), we remove the Adaptive-TM layers from ATM-GCN modules. We conduct ablation study experiments on CAS(ME)$^3$ and the results are reported in Table \ref{tab:ablation}.

\begin{figure*}[ht]
\centering
\includegraphics[width=0.75\textwidth]{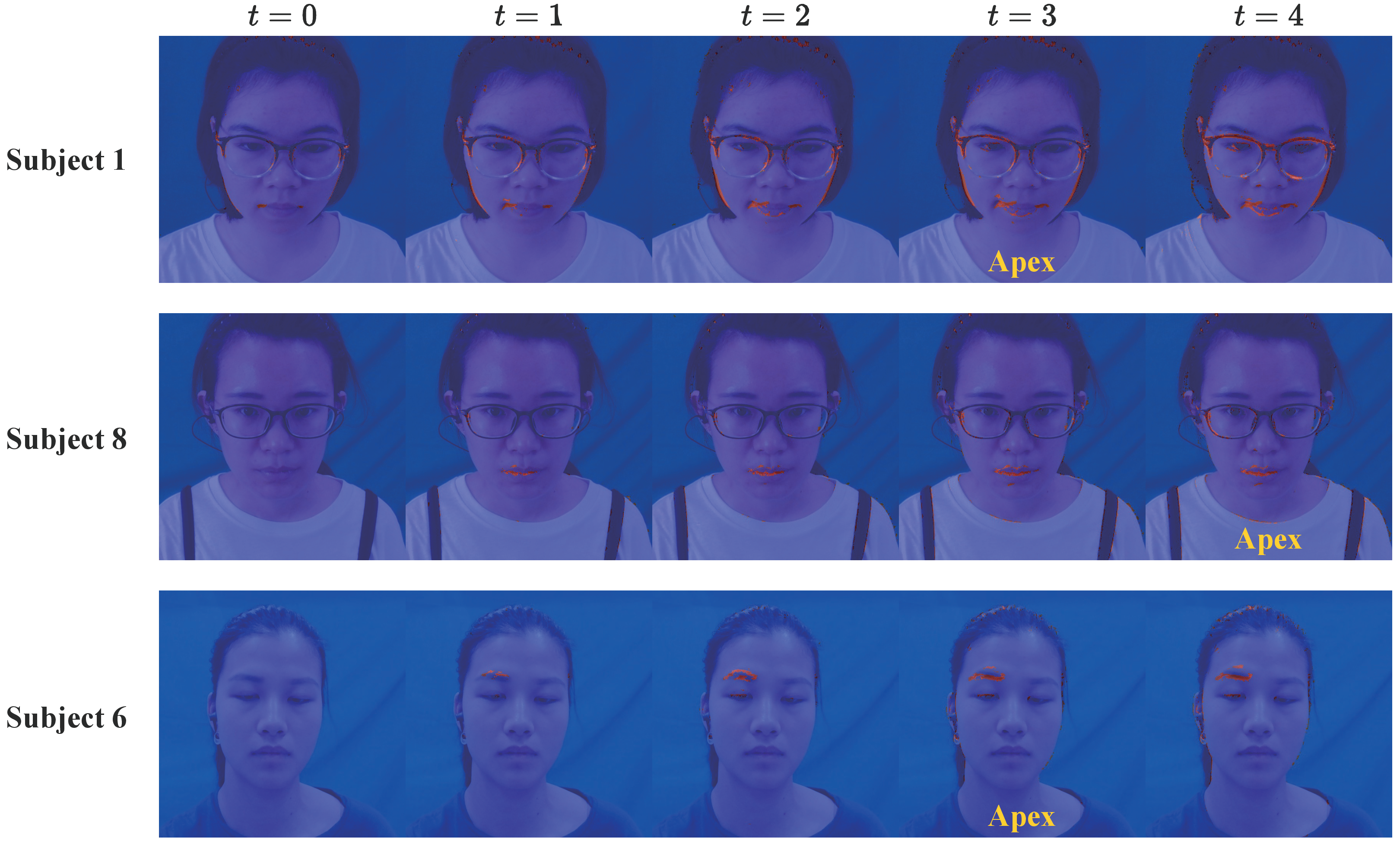} 
\vspace{-8pt}
\caption{Visualization of attention maps of samples from Subject 1,8,6 in CAS(ME)$^3$ respectively. Best viewed in color.}
\label{fig:attention_map}
\end{figure*}

All three variants of ATM-GCN encounter performance drop. Compared to other models, ATM-GCN(\textit{w/o} GCN) cannot achieve comparable results since it fuses all motion features using Pooling layers, which cannot solve temporal redundancy. 
Comparing ATM-GCN(full) with ATM-GCN(\textit{w/o} ATM), we can observe over 2\% performance improvement on both UF1 and UAR, which proves that our proposed Adaptive-TM layers can better adaptively aggregate global and local motion features inherent in micro-expressions.

\subsection{Visualization}

We visualize attention maps of three samples from CAS(ME)$^3$ dataset, as shown in Figure \ref{fig:attention_map}. We use attention maps from the first CA block and visualize 5 of them in each sequence. For each sample, its attention maps are sorted by their temporal locations in the sequence. The regions of micro facial movements are highlighted in red.
As can be seen from Figure \ref{fig:attention_map}, our method manages to capture the micro facial movements in different temporal locations of the sequence. For example, in the given sample of Subject 1, micro movements are highlighted around corners of the mouth. Within the 5 attention maps, our method can successfully focus on the most important facial movements while ignoring other irrelevant ones in each frame, demonstrating that our method can better capture temporal dependencies and further aggregate local motion features for micro-expression recognition.

\section{Conclusion}
In this paper, we introduce a novel framework ATM-GCN for Micro Emotion Recognition primarily driven by motion information. ATM-GCN incorporates temporal dependencies of frame pairs to exploit motion information and manages to pay more attention to the most important motion information from the whole sequence. 
Moreover, our method mitigates redundancy by modeling relations of motion information in different temporal locations.
Experimental results show that ATM-GCN surpasses existing state-of-the-art methods, demonstrating the effectiveness of our proposed method.

\noindent \textbf{Limitations:} Given a long clip, our method will tend to learn dense dependencies between nodes, therefore encountering overfitting problems on small datasets.
We will investigate how to improve robustness on small datasets in future work.

\section*{Acknowledgments}
This work was partially supported by the the National Natural Science Foundation of China (No. 62072462), National Key  R\&D  Program  of  China  (No.2020AAA0108600), and Beijing Natural Science Foundation (No. L233008).

\bibliographystyle{IEEEtran}
\bibliography{ref}

\begin{thebibliography}{10}
\providecommand{\url}[1]{#1}
\csname url@samestyle\endcsname
\providecommand{\newblock}{\relax}
\providecommand{\bibinfo}[2]{#2}
\providecommand{\BIBentrySTDinterwordspacing}{\spaceskip=0pt\relax}
\providecommand{\BIBentryALTinterwordstretchfactor}{4}
\providecommand{\BIBentryALTinterwordspacing}{\spaceskip=\fontdimen2\font plus
\BIBentryALTinterwordstretchfactor\fontdimen3\font minus
  \fontdimen4\font\relax}
\providecommand{\BIBforeignlanguage}[2]{{%
\expandafter\ifx\csname l@#1\endcsname\relax
\typeout{** WARNING: IEEEtran.bst: No hyphenation pattern has been}%
\typeout{** loaded for the language `#1'. Using the pattern for}%
\typeout{** the default language instead.}%
\else
\language=\csname l@#1\endcsname
\fi
#2}}
\providecommand{\BIBdecl}{\relax}
\BIBdecl

\bibitem{Cai2018}
J.~Cai, Z.~Meng, A.~Khan \emph{et~al.}, ``Island loss for learning
  discriminative features in facial expression recognition,'' in \emph{FG},
  2018.

\bibitem{Ding2017}
H.~Ding, S.~Zhou, and R.~Chellappa, ``Facenet2expnet: Regularizing a deep face
  recognition net for expression recognition,'' in \emph{FG}, 2017.

\bibitem{Li2022f}
H.~Li, M.~Sui, Z.~Zhu \emph{et~al.}, ``Mmnet: Muscle motion-guided network for
  micro-expression recognition,'' \emph{arXiv preprint}, 2022.

\bibitem{Zhang2022b}
L.~Zhang, X.~Hong, O.~Arandjelovic, and G.~Zhao, ``Short and long range
  relation based spatio-temporal transformer for micro-expression
  recognition,'' \emph{TAC}, 2022.

\bibitem{Xie2020}
H.~Xie, L.~Lo, H.~Shuai, and W.~Cheng, ``Au-assisted graph attention
  convolutional network for micro-expression recognition,'' in \emph{MM}, 2020.

\bibitem{Kipf2017a}
T.~N. Kipf and M.~Welling, ``Semi-supervised classification with graph
  convolutional networks,'' 2017.

\bibitem{Skianis2018}
K.~Skianis, F.~Malliaros, and M.~Vazirgiannis, ``Fusing document, collection
  and label graph-based representations with word embeddings for text
  classification,'' in \emph{TextGraphs}, 2018.

\bibitem{Davison2018b}
A.~Davison, C.~Lansley \emph{et~al.}, ``Samm: A spontaneous micro-facial
  movement dataset,'' \emph{TAC}, 2018.

\bibitem{Yan2014}
W.~Yan, X.~Li, S.~Wang \emph{et~al.}, ``Casme ii: An improved spontaneous
  micro-expression database and the baseline evaluation,'' \emph{PLOS ONE},
  2014.

\bibitem{See2019}
J.~See, M.~Yap, J.~Li \emph{et~al.}, ``Megc 2019 {\textendash} the second
  facial micro-expressions grand challenge,'' in \emph{FG}, 2019.

\bibitem{Li2013}
X.~Li, T.~Pfister, X.~Huang \emph{et~al.}, ``A spontaneous micro-expression
  database: Inducement, collection and baseline,'' in \emph{FG}, 2013.

\bibitem{Li2022e}
J.~Li, Z.~Dong \emph{et~al.}, ``Cas(me){\textsuperscript{3}}: A third
  generation facial spontaneous micro-expression database with depth
  information and high ecological validity,'' \emph{TPAMI}, 2022.

\bibitem{Nguyen2023}
X.~Nguyen, C.~Duong, X.~Li \emph{et~al.}, ``Micron-bert: Bert-based facial
  micro-expression recognition,'' in \emph{CVPR}, 2023.

\bibitem{Gan2019a}
Y.~Gan, S.~Liong, W.~Yau \emph{et~al.}, ``Off-apexnet on micro-expression
  recognition system,'' \emph{Signal Processing: Image Communication}, 2019.

\bibitem{Liong2019a}
S.~Liong, Y.~Gan, J.~See \emph{et~al.}, ``Shallow triple stream
  three-dimensional cnn for micro-expression recognition,'' in \emph{FG}, 2019.

\bibitem{Xia2020b}
B.~Xia, W.~Wang, S.~Wang \emph{et~al.}, ``Learning from macro-expression: A
  micro-expression recognition framework,'' in \emph{MM}, 2020.

\bibitem{Xia2021a}
B.~Xia and S.~Wang, ``Micro-expression recognition enhanced by macro-expression
  from spatial-temporal domain,'' in \emph{IJCAI}, 2021.

\end{thebibliography}

\end{document}